\documentclass{article}

\usepackage{arxiv}

\usepackage[utf8]{inputenc} % allow utf-8 input
\usepackage[T1]{fo
ntenc}    % use 8-bit T1 fonts
\usepackage{hyperref}       % hyperlinks
\usepackage{url}            % simple URL typesetting
\usepackage{booktabs}       % professional-quality tables
\usepackage{amsfonts}       % blackboard math symbols
\usepackage{nicefrac}       % compact symbols for 1/2, etc.
\usepackage{microtype}      % microtypography
\usepackage{cleveref}       % smart cross-referencing
\Crefname{figure}{Fig.}{Figs.}
\crefname{figure}{Fig.}{Figs.}
\Crefname{table}{Table}{Tables}
\crefname{table}{Table}{Tables}
\Crefname{appendix}{Appendix}{Appendices}
\crefname{appendix}{Appendix}{Appendices}
\usepackage{lipsum}         % Can be removed after putting your text content
\usepackage{graphicx}
\usepackage{doi}

\usepackage{tabularx}
\usepackage{subcaption}
\usepackage{wrapfig}
\usepackage{booktabs}
\usepackage{authblk}
\usepackage{caption}

\author[1,2,*]{\textbf{Patrick Grommelt}}
\author[1,2,*]{\textbf{Louis Weiss}}
\author[1]{\textbf{Franz-Josef Pfreundt}}
\author[2,3]{\textbf{Janis Keuper}}

\affil[1]{Fraunhofer ITWM, Competence Center High Performance Computing Kaiserslautern, Germany}
\affil[2]{Institute of Machine Learning and Analysis (IMLA), Offenburg University, Germany}
\affil[3]{University of Mannheim, Germany}
\affil[ ]{} % Zusätzlicher Platz zwischen Autoren und Institutionen

\affil[*]{\textit{Equal contribution}}

\DeclareCaptionLabelFormat{bold}{\textbf{#1 #2}}
\begin{document}

\title{Fake or JPEG? Revealing Common Biases in Generated Image Detection Datasets} 

\maketitle

\begin{abstract}
	The widespread adoption of generative image models has highlighted the urgent need to detect artificial content, which is a crucial step in combating widespread manipulation and misinformation. Consequently, numerous detectors and associated datasets have emerged. However, many of these datasets inadvertently introduce undesirable biases, thereby impacting the effectiveness and evaluation of detectors.

	In this paper, we emphasize that many datasets for AI-generated image detection contain biases related to \textit{JPEG} compression and image size. Using the \textit{GenImage} dataset, we demonstrate that detectors indeed learn from these undesired factors. Furthermore, we show that removing the named biases substantially increases robustness to \textit{JPEG} compression and significantly alters the cross-generator performance of evaluated detectors. Specifically, it leads to more than 11 percentage points increase in cross-generator performance for \textit{ResNet50} and \textit{Swin-T} detectors on the \textit{GenImage} dataset, achieving state-of-the-art results.
	
	We provide the dataset and source codes of this paper on the anonymous website: \\ \url{https://www.unbiased-genimage.org}
	
\end{abstract}

% keywords can be removed

\section{Introduction}
\label{sec:intro}

In recent years, generative models have improved significantly in creating photorealistic images, making it hard for humans to distinguish between natural and generated images~\cite{lu2023seeing}. Even though these advancements are a significant achievement in many computer vision applications, it does not take much to realize how the misuse of these models can threaten public safety. To tackle this issue, extensive research has been conducted on the development of robust detection mechanisms and techniques for identifying manipulated or synthetic content that is not actively fingerprinted by the generator side.

\textit{Generative Adversarial Networks} (\textit{GANs})~\cite{goodfellow2014generative} have marked a major breakthrough in generating realistic synthetic images. Nevertheless, spectral analysis reveals that images produced by \textit{GANs} inherently contain distinctive artifacts, making them recognizable. Zhang \textit{et al.}~\cite{zhang2019detecting} observed that \textit{GAN}-generated images exhibit periodic, grid-like patterns in their frequency spectrum, a clear deviation from natural image spectra. Furthermore, Durall \textit{et al.}~\cite{durall2020watch} pointed out that \textit{GANs} fail to replicate the spectral distribution of training data, probably due to the inherent transposed convolution operations of these models. This discrepancy between natural and generated images has enabled the research community to develop highly effective, generator-agnostic detection tools, achieving near-perfect accuracy in identifying synthetic images.

However, the landscape of synthetic content generation has evolved rapidly. One significant advancement came with the introduction of \textit{Denoising Diffusion Probabilistic Models} (\textit{DDPM})~\cite{ho2020denoising}, which represented a paradigm shift in generative modelling. \textit{DDPMs} employ a fundamentally different approach compared to \textit{GANs}, utilizing a diffusion process to generate images, which results in a smoother and more realistic appearance. These generative models have also shown better capabilities in approximating the frequency spectra of natural images~\cite{ricker2024detection, corvi2023intriguing, ojha2023universal}, which caused a lot of detectors developed for the identification of \textit{GAN} generated images to perform poorly on these images ~\cite{ricker2024detection}.  To the best of our knowledge, the problem of finding generator-agnostic and robust detection methods for \textit{DDPM}-generated images has not been solved yet. Even though it seems trivial to correctly classify images which are generated by the same model the detector has seen in training, it does not transfer well to other generative models~\cite{ricker2024detection, corvi2023intriguing, zhu2023genimage, zhu2023gendet}. 

To effectively assess and compare the generalization capabilities of detectors, it is imperative to establish a consensus within the community regarding the benchmark to be employed. Ideally, such benchmarks should closely mimic real-world scenarios by being large-scale and encompassing a diverse array of classes and images produced by a range of distinct generators. To address this need, \textit{GenImage}~\cite{zhu2023genimage} introduced an extensive dataset that includes all natural images from the \textit{ImageNet1k} dataset~\cite{5206848}, as well as approximately an equal number of generated images, originating from various generators with differing architectures. Nevertheless, the evaluation of a detector based on the raw \textit{GenImage} Benchmark is not reliable yet since, as we will show, \textit{JPEG} compression and image size biases are used by detectors during the training. 

This paper's primary goal is to raise awareness of \textit{JPEG} compression and image size biases present in most datasets for generated image detection and emphasize the need for heightened scrutiny to ensure that detectors do not inadvertently learn from undesirable variables. Current evaluation benchmarks are somewhat limited in their interpretability, because they do not ensure whether the detection is really based on generation-specific artifacts and thus applicable for real world usage. Consequently, it becomes challenging to assess the effectiveness of various approaches and determine which ideas warrant further research. We firmly believe that identifying and mitigating biases in datasets for generated image detection is crucial, as it lays the foundation for establishing a robust and transparent research environment for generated image detection.

In summary, our main contributions are as follows:\\ \\
\textbullet{} We demonstrate, using the \textit{GenImage} dataset as an example, that many datasets for generated image detection contain \textit{JPEG} and image size biases, which are subsequently used by the detectors during inference. \\ \\
\textbullet{} We show that removing these biases significantly enhances cross-generator performance, achieving state-of-the-art results on \textit{GenImage} and increasing the average accuracy by more than 11 percentage points for \textit{ResNet50} and \textit{Swin-T} detectors. Additionally, detectors become more robust against distortions due to now learning the actual task of detecting generation specific artifacts.

\section{Related Works}
\label{sec:related_works}

\subsection{Detecting Artifically Generated Images}
The problem of detecting AI-generated images is traditionally approached as a binary classification task \cite{durall2020watch, zhang2019detecting, wang2020cnngenerated, wang2023dire, sarkar2023shadows}. However, for practical use in real-world scenarios, detector models not only need to perform well on the training and validation datasets but also exhibit two critical characteristics. Firstly, they must generalize effectively to generative models that were not part of their training data, as the specific generator is typically unknown. This concept is referred to as cross-generator performance. Secondly, these models need to be robust against various transformations such as resizing, compression, or noise, as attackers may attempt to manipulate content to deceive the detector and social media platforms often apply these transformations by default.

\subsubsection*{Detecting GAN Generated Images}
Research has indicated that \textit{GAN}-generated images contain distinctive patterns in the frequency spectrum of their Discrete Cosine Transformation~\cite{zhang2019detecting}. These patterns arise due to the presence of upsampling layers~\cite{durall2020watch}, making \textit{GAN}-generated content detectable. Wang \textit{et al.}~\cite{wang2020cnngenerated} have demonstrated that even a basic CNN classifier, \textit{ResNet50}~\cite{he2015deep}, can successfully identify images generated by specific \textit{GANs}. They have also shown that augmentations like random \textit{JPEG} compression or Gaussian Noise significantly enhance cross-generator performance. To encourage the CNN classifier to focus on the frequency information, which is a generator-agnostic characteristic, Gragnaniello \textit{et al.}~\cite{gragnaniello2021gan} suggest removing the initial downsampling layers of \textit{ResNet50}. Cozzolino \textit{et al.}~\cite{cozzolino2021universal} have combined these concepts but opted for cropping instead of resizing during training. This approach prevents the loss of frequency information and enhances robustness against resizing augmentations during inference. These detectors are often referred to as universal \textit{GAN} detectors due to their impressive ability to perform well on unseen \textit{GANs} and their robustness against various transformations, making them suitable for real-world scenarios.

\subsubsection*{Detecting Diffusion Model Generated Images}
Unfortunately, several existing studies~\cite{ricker2024detection, corvi2023intriguing, zhu2023genimage} have indicated that universal \textit{GAN} detectors do not effectively classify images generated by \textit{Diffusion Models} (\textit{DM}). While the frequency spectrum of \textit{DM}-generated images still exhibits differences compared to natural images, these artifacts differ from those present in \textit{GAN}-generated images. Furthermore, these artifacts tend to be more specific to the particular \textit{Diffusion Model} generator, making the achievement of good cross-generator performance a considerably more challenging task.
It is also worth noting that \textit{Diffusion Models} are capable of replicating transformation artifacts found in the images they have been trained on with great accuracy~\cite{corvi2023intriguing}. For instance, if the \textit{DM} generator has been exposed to \textit{JPEG}-compressed images during training, the generated images will contain artifacts that are similar to \textit{JPEG}-artifacts. Research also indicates that transformation artifacts in frequency space can closely resemble those generated by \textit{Diffusion Models}. This complicates the task of ensuring robustness to transformations and as our research will demonstrate, it underscores the importance of carefully selecting the training dataset for the detection model.

Nevertheless, it still appears relatively easy to detect images generated by a specific generator, even with small detector models and training datasets~\cite{zhu2023genimage, sinitsa2023deep}. Prior work has shown that \textit{Diffusion Models}, for example, exhibit systematic errors in projective geometry~\cite{sarkar2023shadows}. However, to the best of our knowledge, no detection model has achieved good cross-generator performance and robustness in realistic application scenarios.Promising methods have emerged that achieve outstanding results on small datasets, but they have yet to prove their mettle on larger and more diverse benchmarks like the \textit{GenImage} dataset~\cite{zhu2023genimage}. \textit{DIRE}~\cite{wang2023dire}, for instance, measures the disparity between an image and its reconstructed version of a \textit{Diffusion Model} to detect images from both \textit{Diffusion Models} and \textit{GANs}. They report nearly perfect results in cross-generator performance and robustness on the \textit{DIRE} dataset, but performance drastically declines when evaluated on \textit{GenImage}~\cite{zhu2023gendet}. \textit{GenDet}~\cite{zhu2023gendet} significantly improves results by approaching the task as an outlier detection problem instead of binary classification. Effectively, they train a Teacher/Student-network and, to the best of our knowledge, present state-of-the-art results on \textit{GenImage}.

\subsection{GenImage Dataset}
The \textit{GenImage} dataset~\cite{zhu2023genimage}, stands out as one of the biggest and most diverse datasets for generated image detection. Its primary goal is to establish a unified benchmark within the research community, facilitating the evaluation of detection methods on a standardized dataset.
Built upon the \textit{ImageNet} dataset~\cite{5206848}, \textit{GenImage} incorporates the original \textit{ImageNet} images as the natural-image class. For the AI-generated class, \textit{GenImage} includes images generated by eight generative models, comprising seven \textit{Diffusion Models} and one \textit{GAN}. These models are specifically \textit{Midjourney} (MJ)~\cite{midjourney2022}, \textit{Stable Diffusion V1.5} (SD1.5) ~\cite{autoamtic1111-stable-diffusion-webui}, \textit{Stable Diffusion V1.4} (SD1.4)~\cite{autoamtic1111-stable-diffusion-webui}, \textit{Wukong}~\cite{wukong_2022}, \textit{VQDM}~\cite{gu2022vector}, \textit{ADM}~\cite{dhariwal2021diffusion}, \textit{GLIDE}~\cite{nichol2022glide}, and \textit{BigGAN}~\cite{brock2019large}. They are either Text-to-Image \textit{(MJ, SD, Wukong, VQDM, GLIDE)}, utilizing \textit{ImageNet}-classes as text prompts, or class-conditional \textit{(ADM, BigGAN)}, to guarantee a consistent content distribution between natural and AI-generated images.
The dataset is organized into eight distinct subsets, one for every generative model. Each subset contains training and validation data, with a nearly equal number of natural images from \textit{ImageNet} and generated images from the respective generative model. To assess the performance of a detection method, the standard approach involves training on one \textit{GenImage} subset and evaluating on the others. This method enables the measurement of a detection method's cross-generator performance, providing insights into its ability to generalize across different generative models. The evaluation process is quantified by a cross-generator matrix. \cref{fig:reproduced} illustrates the results presented in the original paper using a basic \textit{ResNet50} classifier, with accuracy (in \%) serving as the key metric.

\section{Common Biases in Datasets for AI-Generated Image Detection}
When examining various datasets used for AI-generated image detection, it is evident that a significant disparity exists in terms of compression techniques. As demonstrated in \cref{tab:datasets}, a common practice involves storing all artificially AI-generated images within the dataset as \textit{PNG} files, which involves lossless compression, while natural images are stored in \textit{JPEG} format, involving lossy compression and introducing noticeable artifacts. %It's worth noting that even if random \textit{JPEG} augmentation sampled from the same distribution for all images is employed during training, as common in practice \cite{wang2020cnngenerated, ojha2023universal}, natural images will have in average a higher level of \textit{JPEG} compression due to their initial compression and the idempotent nature of the \textit{JPEG} transformation.
This prompts us to investigate whether detectors trained on such datasets rely on detecting \textit{JPEG} compression artifacts for classification.

\begin{table}[hb]
    \caption{Overview over common AI-Generated Image Detection Datasets and their image compression and size properties.}
    \centering
    \newcolumntype{L}{>{\centering\arraybackslash\fontsize{8pt}{8pt}\selectfont}X}
    %\newcolumntype{Y}[1]{>{\raggedright\arraybackslash}p{#1}} %Linksbündig
    \newcolumntype{Y}[1]{>{\centering\arraybackslash\fontsize{8pt}{8pt}\selectfont}p{#1}}
    \begin{tabularx}{\textwidth}{|L||Y{1.7cm}|L|L||Y{1.7cm}|L|L|}

    \hline
    Dataset & \multicolumn{3}{c||}{Natural Images} & \multicolumn{3}{c|}{Synthetic Images} \\
    \hline
     & Compression & Size & Source & Compression & Size & Source \\
    \hline
    \hline
    \textit{GenImage} ~\cite{zhu2023genimage}& \textit{JPEG} & diverse & \textit{ImageNet} & \textit{PNG} & 128x128, 256x256, 512x512, 1024x1024 & \textit{LAION}, \textit{ImageNet}, unknown \\
    \hline
    Wang \textit{et al.}~\cite{wang2020cnngenerated} & mostly \textit{JPEG} (saved as \textit{PNG}) & diverse (resized to 256x256) & \textit{LSUN}, \textit{ImageNet}, \textit{CelebA}, \textit{COCO} & \textit{PNG} & 256x256 & trained on same data as natural images \\
    \hline
    \textit{DIRE}~\cite{wang2023dire} & \textit{JPEG} & diverse & \textit{LSUN}, \textit{CelebA-HQ}, \textit{ImageNet} & \textit{PNG} & 256x256, 512x512, 1024x1024 & \textit{LSUN}, \textit{ImageNet}, unknown \\
    \hline
    Epstein \textit{et al.}~\cite{epstein2023online} & \textit{JPEG} & diverse & \textit{LAION} & \textit{PNG} & 256x256, 512x512, 1024x1024 & \textit{LSUN}, \textit{ImageNet}, \textit{LAION}, unknown \\
    \hline
    Ricker \textit{et al.}~\cite{ricker2024detection} & \textit{JPEG} & 256x256 (resized to) & \textit{LSUN-Bedroom} & \textit{PNG} & 256x256 & \textit{LSUN-Bedroom} \\
    \hline
    Ojha \textit{et al.}~\cite{ojha2023universal} & \textit{JPEG} & diverse & \textit{ImageNet}, \textit{LAION} & \textit{PNG} & 256x256 & \textit{ImageNet}, unknown \\
    \hline
    \end{tabularx}
    \label{tab:datasets}
\end{table}

Furthermore, considering that the generators employed for most of AI-generated image detection datasets generate images of a fixed size, as opposed to the diverse size distribution found in natural images, we investigated whether this disparity in size distribution could potentially cause detectors to distinguish between natural and generated images based on their dimensions.\\
The reason why this is problematic is twofold: firstly, it can lead to an overestimation of the performance in detecting generation-specific artifacts, since evaluation data often contains the same biases from which the detector learned during training. Secondly, detectors having such a reliance on undesirable variables do not generalize well to datasets where biases related to these variables are absent, which is the case in real world scenarios. This lack of robustness hinders the adaptability of detectors to changes in these variables.

\begin{figure}
    \centering
    \begin{subfigure}{0.45\textwidth}
        \centering
        \includegraphics[width=\linewidth]{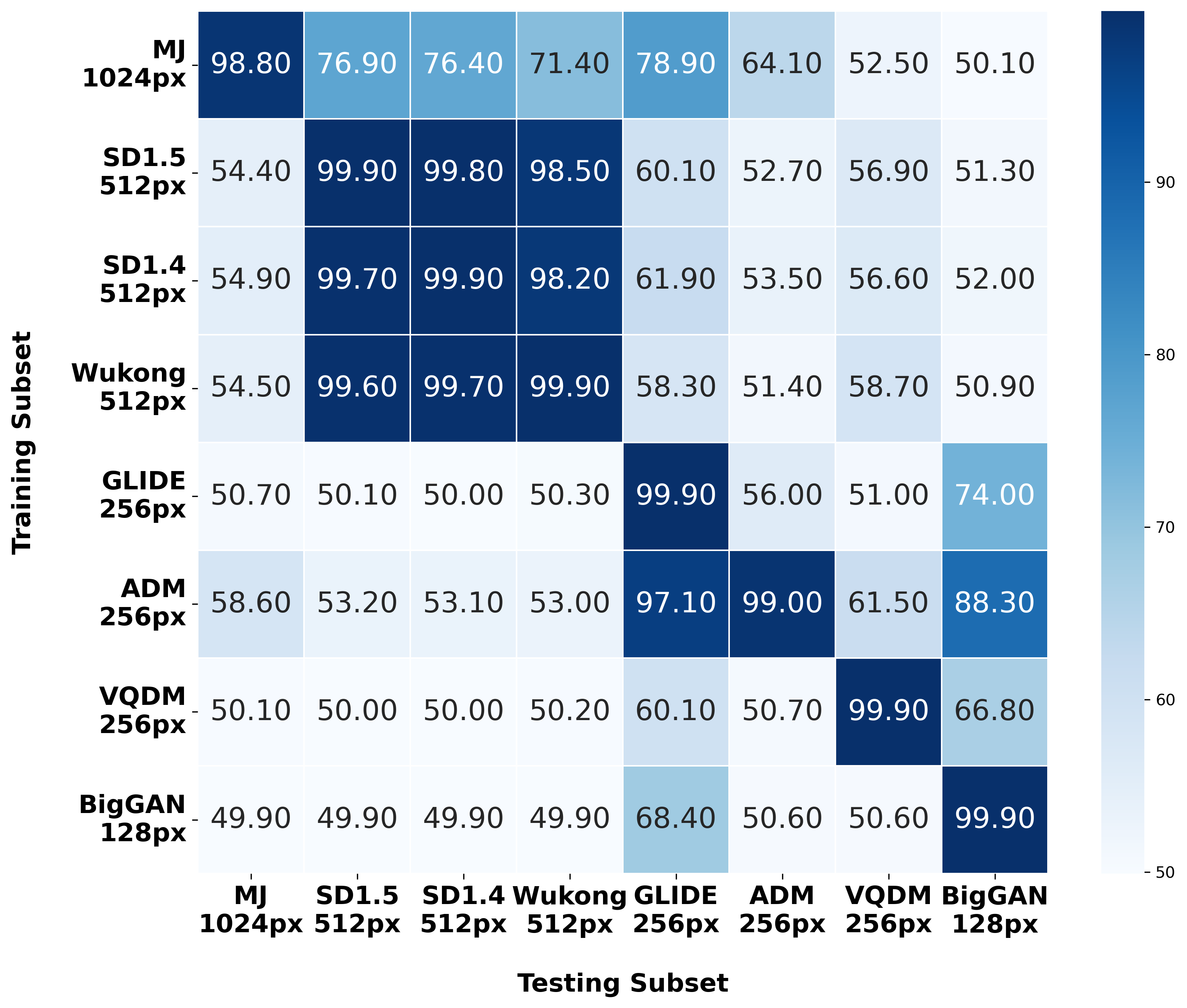}
        \caption{\textit{GenImage} Results}
        \label{fig:genimage}
    \end{subfigure}
    \hfill
    \begin{subfigure}{0.45\textwidth}
        \centering
        \includegraphics[width=\linewidth]{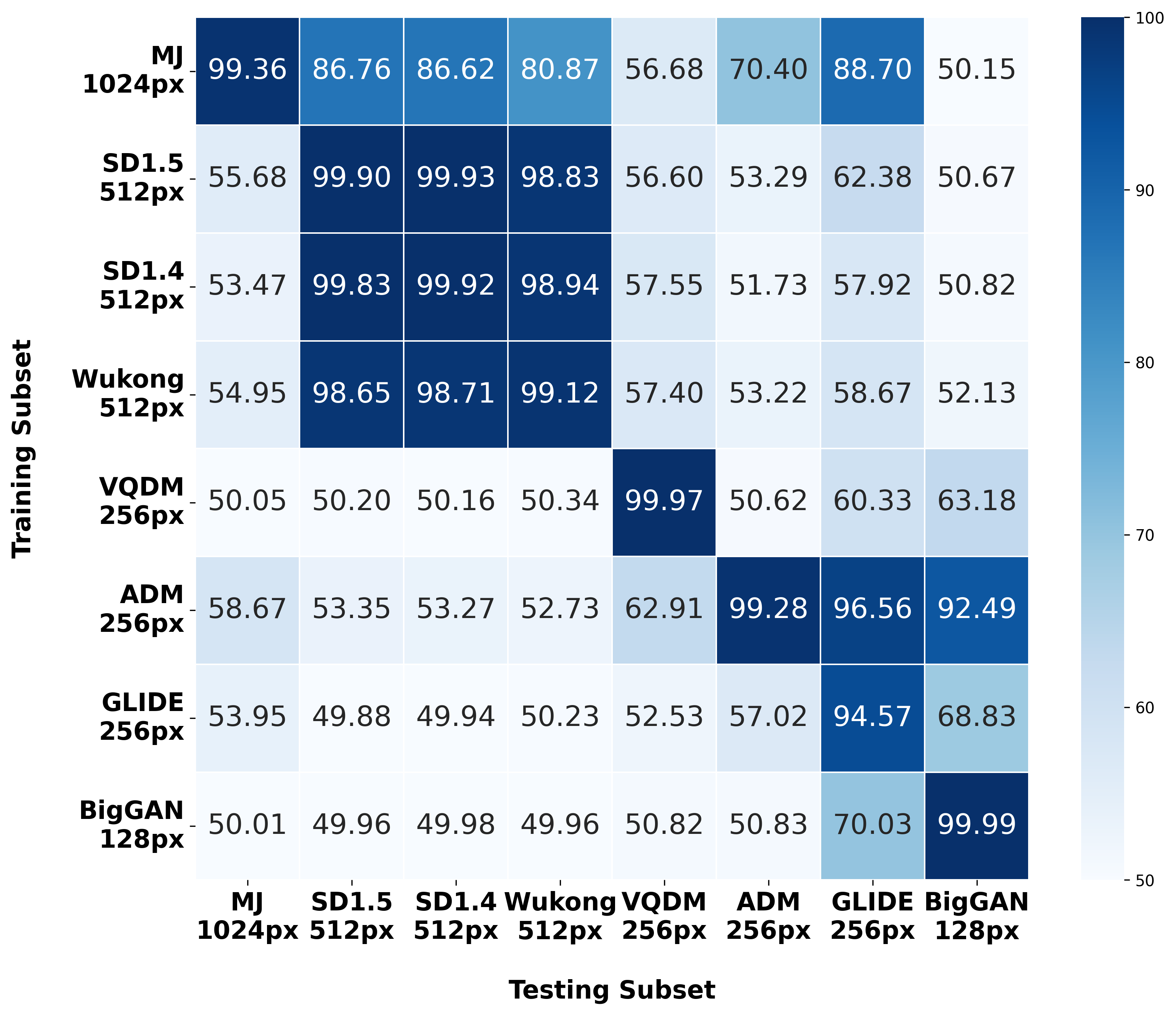}
        \caption{Reproduced Results}
        \label{fig:genimagereproduced}
    \end{subfigure}
    \caption{Reproduced Results. Cross-generator performance of a \textit{ResNet50} classifier from the \textit{GenImage} paper (left), and our reproduced results (right). The matrix shows the accuracy (in \%) of a model trained on a \textit{GenImage} training subset (row), when evaluated on a \textit{GenImage} validation subset (column).}
    \label{fig:reproduced}
\end{figure}
We conducted experiments using the \textit{GenImage} dataset to assess whether detectors, trained on datasets containing such biases, inadvertently acquire information from these undesirable variables. For these experiments, we trained \textit{ResNet50} detectors since this is the baseline methodology in the \textit{GenImage} paper. We used the provided code from \textit{GenImage} and successfully replicated the reported results, as depicted in \cref{fig:reproduced}. The results show near perfect accuracy when the detectors are tested on the subset containing generated images from the same generative model they were trained on, but poor accuracy on others. %However, the cross-generator performance is very poor and mostly around 50\%. It is important to note that almost all errors are False-Negatives, which is expected since the natural images in all \textit{GenImage} subsets are sampled from the same dataset, namely \textit{ImageNet}. Exceptionally, some cases demonstrate effective generalization, yet the reasons for such variance in generalization across different Diffusion Models remain an unresolved issue within the research community.
By sorting the columns and rows according to the output size of the respective generative models, as we have done, it becomes apparent that better generalization tends to occur near the diagonal. This observation raises the question of whether similarity in generator size may enhance cross-generator performance, a topic we discuss in \cref{sec:3.2} and \cref{sec:4.2}. 

\label{headings}

\subsection{JPEG Compression Bias} \label{sec:3.1}
% Is this small paragraph necessary? Or can we get rid of it and just start with the next one?
As mentioned, the natural images used by \textit{GenImage} are sourced from the \textit{ImageNet} dataset. \hyperref[app:a]{Appendix A} illustrates the distribution of \textit{JPEG} quality factors employed in \textit{ImageNet} on a logarithmic scale, with the majority of images compressed using a quality factor of 96~\cite{compression-imagenet}. On the other hand, generated images from \textit{GenImage} are uncompressed. This clear disparity of compression between natural and generated images is common in many datasets, as shown by \cref{tab:datasets}.

To investigate whether detectors trained on datasets containing such compression disparities partially function as \textit{JPEG} detectors, we conducted two experiments on the \textit{GenImage} dataset. Initially, we examined whether compressing the dataset's generated images influences their classification as natural. To do so, we used the \textit{ResNet50} detector trained on raw \textit{GenImage} dataset (\textit{cf.} \cref{fig:reproduced}) and evaluated the cross-generator performance of these detectors on test data that was progressively compressed with lower quality factors.

\begin{figure}
    \centering
    \includegraphics[width=1\linewidth]{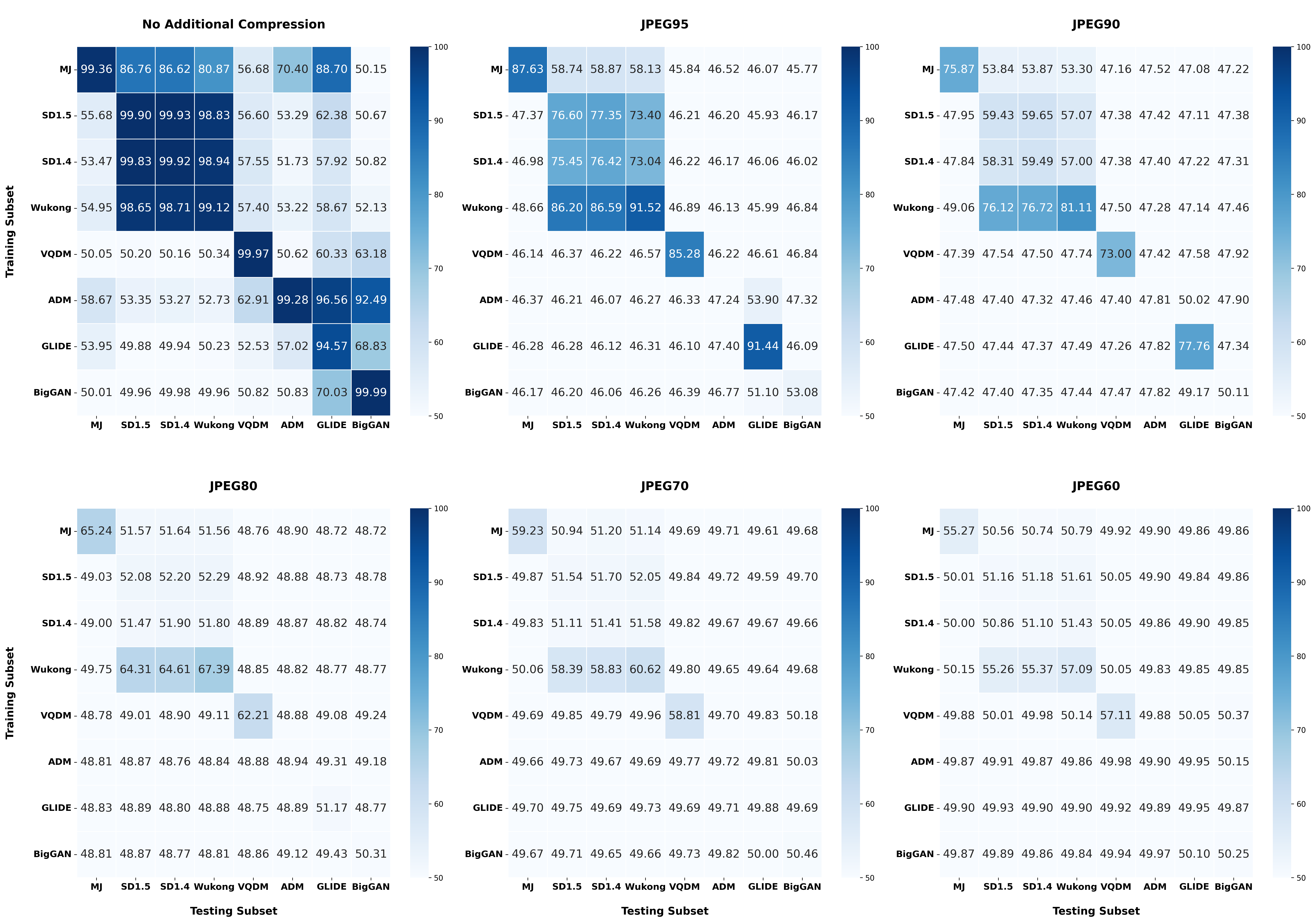}
    \caption{Cross-generator performance of detectors trained on raw \textit{GenImage} for different compression quality factors, given in accuracy (in \%).}
    \label{fig:jpegcompression}
\end{figure}

\cref{fig:jpegcompression} illustrates a strong decline in accuracy with increased compression, even for high quality factors like 95. Analyzing the confusion-matrix of the results revealed that the precision in detecting AI-generated images consistently remained close to one, whereas the recall dropped significantly (\textit{cf.} \hyperref[app:b]{App. B}). This suggests that compressing a generated image considerably increases the likelihood of the model classifying it as natural.

However, this experiment alone does not confirm that the detector learned to identify \textit{JPEG} compression, as compression might destroy generation-specific artifacts. To address this possibility, we used uncompressed natural PNG images from the FFHQ dataset~\cite{karras2019stylebased} to observe how compression impacts their classification. Utilizing the same detectors, we evaluated their performance on the FFHQ images as they were \textit{JPEG}-compressed with different quality factors.

\cref{tab:ffhq} shows the results for the detector trained on the \textit{Midjourney} subset of \textit{GenImage}. 
We observe that the detectors' ability to accurately classify a natural image improves as the level of compression increases. Specifically, the detector's accuracy improves from 80.4\% for uncompressed \textit{PNG} images up to 100\% for natural images compressed with a low-quality factor of 60. Even a small compression with qualtity factor 95 leads to a much better accuracy of 94.8\%. This experience underscores the direct influence of compression towards the classification as natural, as compression, in this scenario, can´t result in the potential destruction of generation-specific artifacts.

\subsection{Size Distribution Bias}\label{sec:3.2}

The generated images in the \textit{GenImage} dataset originate from eight different generators, producing images of four different sizes: 1024×1024 (\textit{MJ}), 512×512 (\textit{SD4}, \textit{SD5}, \textit{Wukong}), 256×256 (\textit{GLIDE}, \textit{ADM}, \textit{VQDM}), and 128×128 (\textit{BigGAN}). In contrast, the natural images sourced from \textit{ImageNet} contain images of various sizes, as shown in \hyperref[app:a]{Appendix A}.

\begin{table}[t]
\newcolumntype{C}[1]{>{\centering\arraybackslash}p{#1}} % Zentrierte Spalten mit fester Breite
\newcolumntype{L}[1]{>{\raggedright\arraybackslash}p{#1}} % Links ausgerichtete Spalten mit fester Breite
\centering
\caption{Accuracy (in \%) of the \textit{ResNet50} detector trained on the \textit{Midjourney} subset on 1024×1024 natural FFHQ images when progressively increasing compression.}
\label{tab:ffhq}
\begin{tabular}{L{2.5cm}C{1.2cm}C{1.2cm}C{1.2cm}C{1.2cm}C{1.2cm}C{1.2cm}}
\toprule
\textbf{Compression} & \textit{PNG} & \textit{JPEG95}
 & \textit{JPEG90} & \textit{JPEG80} & \textit{JPEG70} & \textit{JPEG60} \\ 
\midrule
\textbf{Accuracy} & 80.45 & 94.84 & 98.93 & 99.95 & 99.99 & 100.0 \\
\bottomrule
\end{tabular}

\end{table}

We anticipated that this fundamental difference in size distribution could be exploited by some detectors to discern whether an image is generated or not. For instance, consider a \textit{ResNet50} detector which initially resizes all the input data to 224×224 pixels, as the \textit{ResNet50} detector evaluated in \textit{GenImage}, and is trained with generated images from the \textit{GLIDE} generator. Given that most natural images from the \textit{GenImage} dataset have dimensions around 450×450 pixels, they would typically undergo more resizing than the 256×256 images from \textit{GLIDE}. Consequently, the detector could potentially extract information about the nature of an image by detecting the strength of resizing artifacts. Furthermore, for non-square images, resizing results in significant information loss along one axis, leading to frequency artifacts. However, this problem extends beyond detection methods that rely solely on resizing as a preprocessing step. Detectors employing cropping to achieve a uniform input size are still subject to bias in terms of object size. Discrete-Cosine-Transormation based detectors will contain similar biases as well, as larger images will have more data points (pixels), which affects the frequency spectrum towards having higher frequencies. 

\begin{figure}[t!]
    \centering
    \includegraphics[width=1\linewidth]{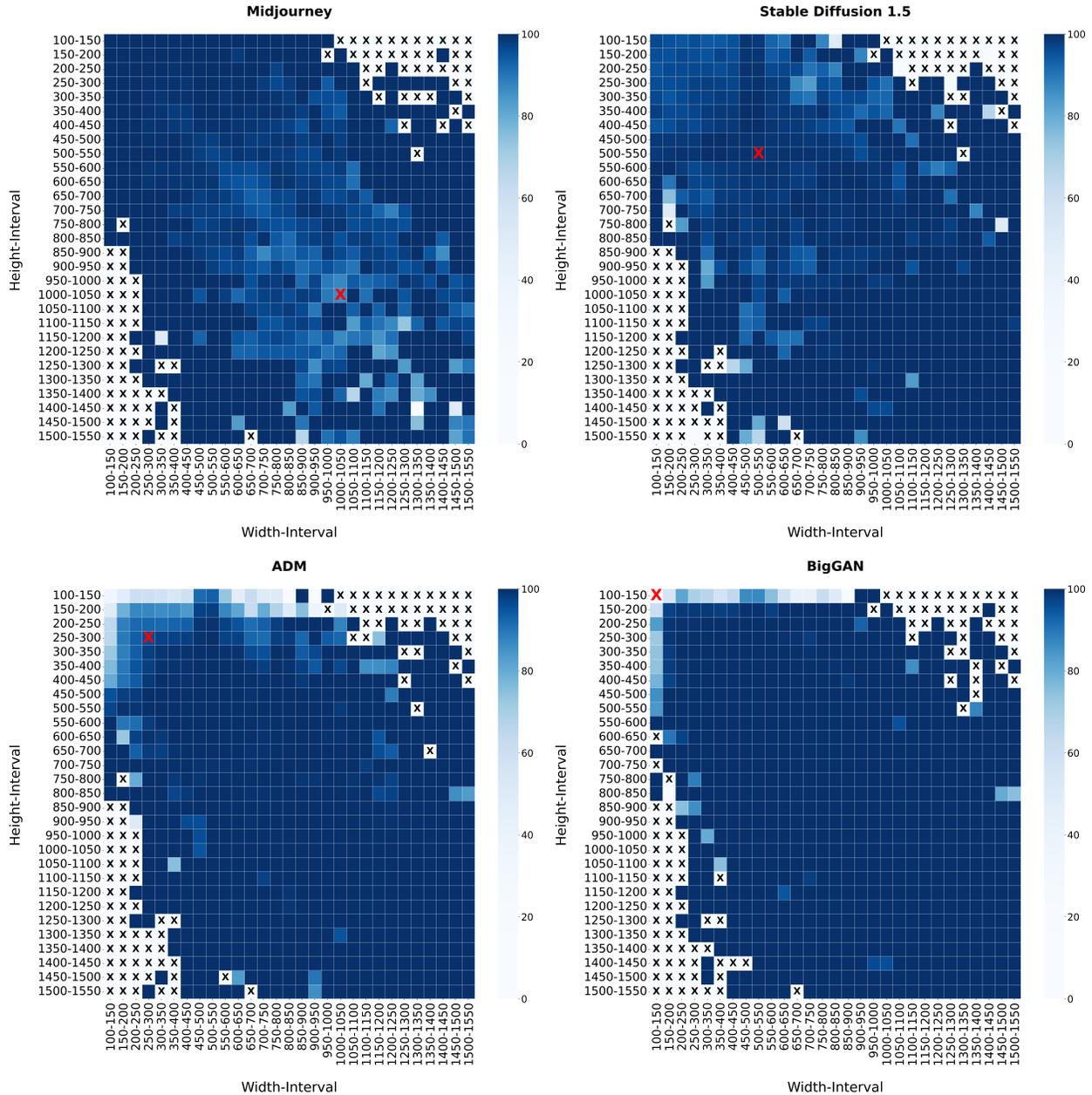}
    \caption{Accuracy (in \%) of detectors trained on different \textit{GenImage} subsets when evaluated on natural images of varying sizes. Black crosses indicate intervals without any data. The red cross marks the size interval of generated images corresponding to this subset.}
    \label{fig:size_plot}
\end{figure}

To investigate whether detectors trained on the \textit{GenImage} dataset indeed acquire information about image size, we conducted an experiment to evaluate how well a detector performs on natural images of various sizes. A decrease in performance for natural images that closely match the dimensions of the generated images the detector was trained on might suggest that the detector is, to some extent, a size detector. \cref{fig:size_plot} displays the accuracy of \textit{ResNet50} detectors on natural images across different size intervals. For clarity, we show the performance of detectors trained on one subset for each generator-size available in the \textit{GenImage} dataset. Each diagram illustrates the performance of a detector trained on a specific \textit{GenImage} subset. %with a red cross marking the size interval of the generated images encountered during training.
The analysis included all natural \textit{ImageNet} images not seen by the detector during training, effectively using the training as well as validation data from other \textit{GenImage} subsets. \hyperref[app:a]{Appendix A} illustrates the number of \textit{ImageNet} images in each interval. It is important to highlight that some intervals contain very few images, which is why the single cells are not as informative as the global trend. %The black crosses mark intervals without any images.
Note that we used the models trained in \cref{sec:4.1}, since otherwise the classification of natural images is highly affected by \textit{JPEG}-artifacts and thus the prediction is near perfect for all intervals.

Our findings indicate a correlation between decreased detector performance and natural images sized similarly to those of the generated images the detector was trained on. These results demonstrate that detectors indeed perform better on natural images that have significantly different sizes compared to the size of the generated images used during detector training. For instance, the \textit{BigGAN} detector accurately classifies most natural images, except from very small images with one side between 100-150, as \textit{BigGAN} images are of size 128×128. \textit{ADM} images are of size 256×256, which is why the effect is also visible for slightly bigger images. Conversely, the \textit{Midjourney} detector shows reduced performance on larger natural images, given that \textit{Midjourney} images are much bigger than most natural images in \textit{ImageNet}. For \textit{Stable Diffusion}, no significant trend emerges, as the generated images are of similar size as most of the natural images. This detector has to learn more from other discriminative patterns in the interval of the generated images, which does not mean that the the detector did not learn from the bias to classify images outside of the interval.

\section{Removing The Biases}

We have emphasized that disparities in size distribution and compression between natural and generated data can cause detectors to learn from differences that may not be pertinent in real-world contexts. Consequently, we retrained the same detectors using a constrained dataset and reevaluated them to investigate whether the removal of compression and size biases would affect the final evaluation of detectors. Our findings reveal significantly divergent results and notably enhanced cross-generator performance and robustness.

\subsection{JPEG Constraint}
\label{sec:4.1}
\begin{figure}[h]
    \centering
    \includegraphics[width=1\linewidth]{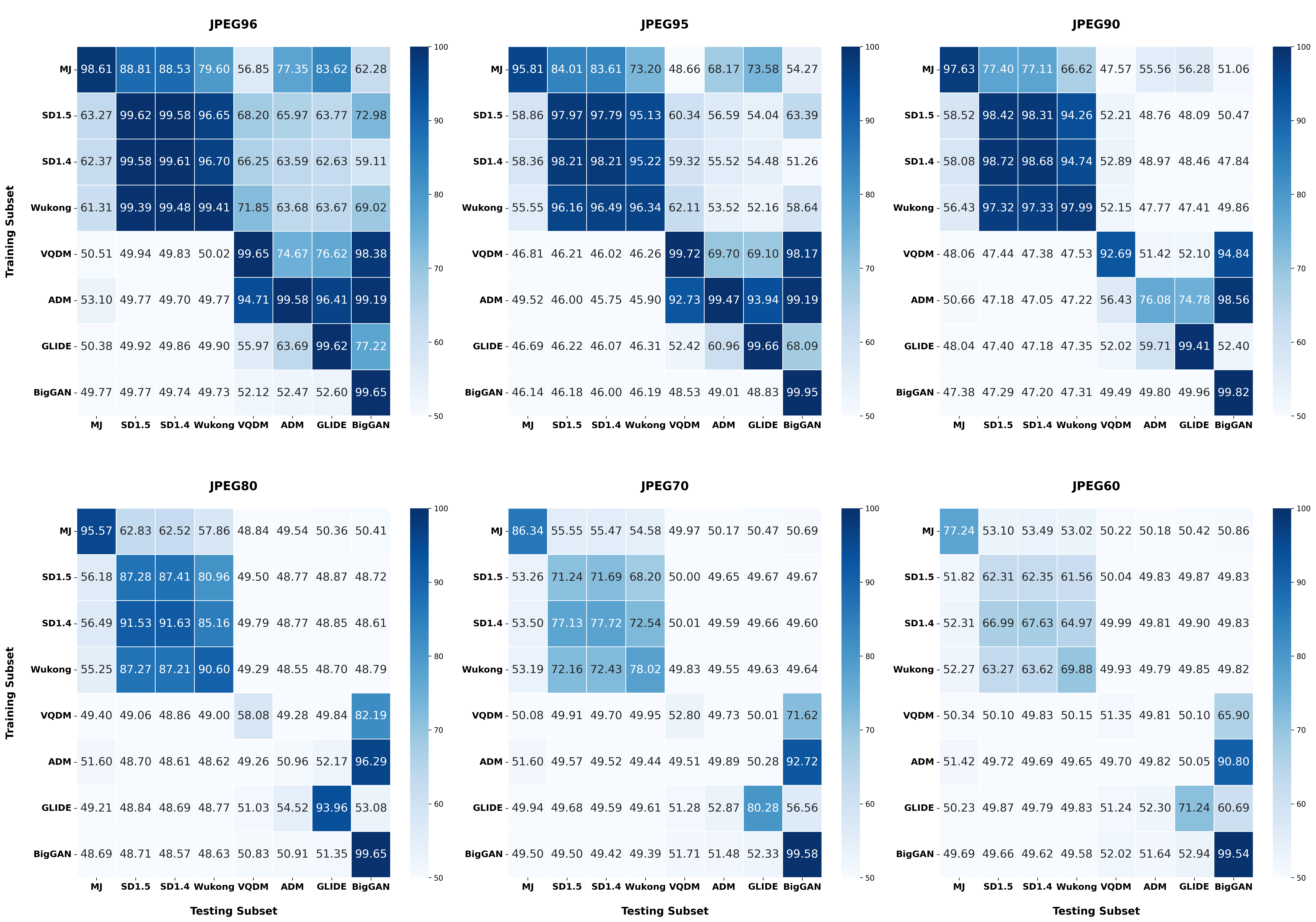}
    \caption{Cross-generator performance given in accuracy (in \%) of detectors trained only with images compressed with \textit{JPEG} quality factor of 96 from \textit{GenImage} for different Compression Rates.}
    \label{fig:jpegconstraintcompression}
    \setlength{\belowcaptionskip}{-30pt}
\end{figure}

\newcolumntype{C}[1]{>{\centering\arraybackslash}p{#1}} % Neuer Spaltentyp 'C' für zentrierten Text in einer bestimmten Breite
To ensure that the detector does not learn to differentiate between natural and generated data based on \textit{JPEG} compression artifacts, we constrained the quality factor of all images used in training. Specifically, we constructed a training set by exclusively selecting natural images compressed with a quality factor of 96. We then selected an equivalent number of generated images and compressed them using the same quality factor. Subsequently, we trained a \textit{ResNet50} detector using this constrained dataset and reevaluated it's cross-generator and robustness performance. This approach is expected to make the detector's classification less sensitive to \textit{JPEG} compression than before.
\begin{table}[t]
    \centering
    \caption{Comparison of robustness to compression between training on raw \textit{GenImage} and only \textit{JPEG96 GenImage}: Accuracy in \% averaged cross the generator matrix.}
    \label{tab:jpegcompdiff}
    \small
    \begin{tabular}{|l|*{3}{C{3cm}|}}
    \hline & \multicolumn{2}{c|}{\textbf{Training dataset}} & \\
        \hline
        \textbf{Compression} & \textbf{Classic \textit{GenImage}} & \textbf{\textit{JPEG96} (ours)} & \textbf{Difference} \\
        \hline
        \textit{JPEG95} & 53.91 & 67.17 & +13.26 \\
        %\hline
        %\textit{JPEG90} & 51.67 & 63.19 &  +11.52 \\
        \hline
        \textit{JPEG80} & 50.62 & 59.37 &  +8.75 \\
        %\hline
        %\textit{JPEG70} & 50.74 & 56.72 & +5.98 \\
        \hline
        \textit{JPEG60} & 50.58 & 55.07 &  +4.49 \\
        \hline
    \end{tabular}
    \setlength{\belowcaptionskip}{-30pt}
\end{table}
\cref{fig:jpegconstraintcompression} illustrates the cross-generator performance of these detectors when the test data is compressed with increasing quality factors. Unlike the training data, the test data is not subject to any constraints. It is important to note that for \textit{JPEG96}\footnote{We mean with JPEGX a JPEG compression with a quality factor of X.}, compression was applied solely to the generated images, as the majority of natural images is already compressed to \textit{JPEG96} or a lower quality. Consequently, for compression rates other than \textit{JPEG96}, natural images undergo a second round of compression. 
%Interestingly, this doubled compression seems to significantly impact the compression artifacts. We observed this when compressing natural images in training as well: The detectors almost exclusively learned to distinguish whether the images are compressed once or twice.   

Comparing the results shown in \cref{fig:jpegconstraintcompression} with those in \cref{fig:jpegcompression}, which presents the same evaluation for detectors trained on the raw \textit{GenImage} dataset, we observe a strong enhancement in robustness against \textit{JPEG} compression. Specifically, as illustrated in \cref{tab:jpegcompdiff}, we observe an overall improvement of 13.26 accuracy points to \textit{JPEG95} compression, 8.75 for \textit{JPEG80}, and 4.49 for \textit{JPEG60}. For example, the \textit{BigGAN} detectors performance remains nearly intact up to \textit{JPEG60}, whereas previously, the detector incorrectly classified all \textit{JPEG95} images as natural.

These results prove our assumption that biases in compression lead to the detectors learning wrong causalities and not being robust to changes in compression. Previous research \cite{wang2020cnngenerated} has already indicated that introducing random \textit{JPEG} augmentation into the training set enhances detector robustness and generalization to other generators. We firmly believe that this improvement does not solely stem from increasing the training set’s variance but also partially from reducing the bias arising from differences in \textit{JPEG} compression during training. Nevertheless, random-augmentation does not completely remove the bias, since it merely shifts the distributions of quality factors to have a bigger overlap instead of equalizing them. Moreover, when using augmentation, already compressed images are compressed a second time, while uncompressed images are only compressed once. This introduces a bias, as \textit{JPEG} compression is not an idempotent transformation.

Surprisingly, mitigating the \textit{JPEG} bias marginally improves the generalization across different generators. This contradicts our initial expectation that the bias, being present in all \textit{GenImage} subsets, should facilitate transferability. To explain this we propose two hypothesis: Firstly, a training dataset with reduced bias compels the detectors to focus on learning from generation-specific artifacts. These artifacts are not only more resilient to compression but also to other transformations like resizing, thereby enhancing generalization to generators of various sizes. Secondly, research indicates that \textit{Diffusion Models}, when trained on \textit{JPEG}-compressed images, produce artifacts that mimic those of compression artifacts~\cite{corvi2023intriguing}. This suggests that biased detectors do not actually distinguish between compressed and uncompressed images, but rather between genuine compression artifacts and their approximations. This distinction may generalize less effectively to other generative models.

\subsection{Size Constraint}\label{sec:4.2}

To minimize the detector’s reliance on variations in size, we aimed to control this variable by selecting natural images within a specific size range for training, such that the natural images in training are of similar size as those of the corresponding generator. As this selection drastically reduces the number of training samples, we only executed the experiment on subsets with generated images of dimension 512×512, because most natural \textit{GenImage} images have both height and width within the range [450, 550]. 
In order to obtain sufficient training data, we also utilized the natural training data from all \textit{GenImage} training subsets in this interval. Furthermore we still only used \textit{JPEG96} images to mitigate the compression bias. We then sampled the same number of generated images for each 512×512 generator. To avoid disparities in content distribution between natural and generated images, we ensured an equal number of natural and generated images per \textit{ImageNet} class. This data selection in total reduced the number of training samples from approximately 320,000 to 75,000. For the evaluation, we utilized the unconstrained validation sets, to maintain consistency with other experiments. 
To guarantee that all images images contain the same amount of resize artifacts during training, we center-cropped images to 450 - the lower bound of the size interval - before resizing to 224 the final inputsize of the \textit{ResNet50} detectors. During inference, for images of any size, we first resize the input to 512 to allow cropping to 450 and then resizing.

\begin{figure}[t!]
    \centering
    \begin{subfigure}[t]{0.45\textwidth}
        \centering
        \includegraphics[width=\linewidth]{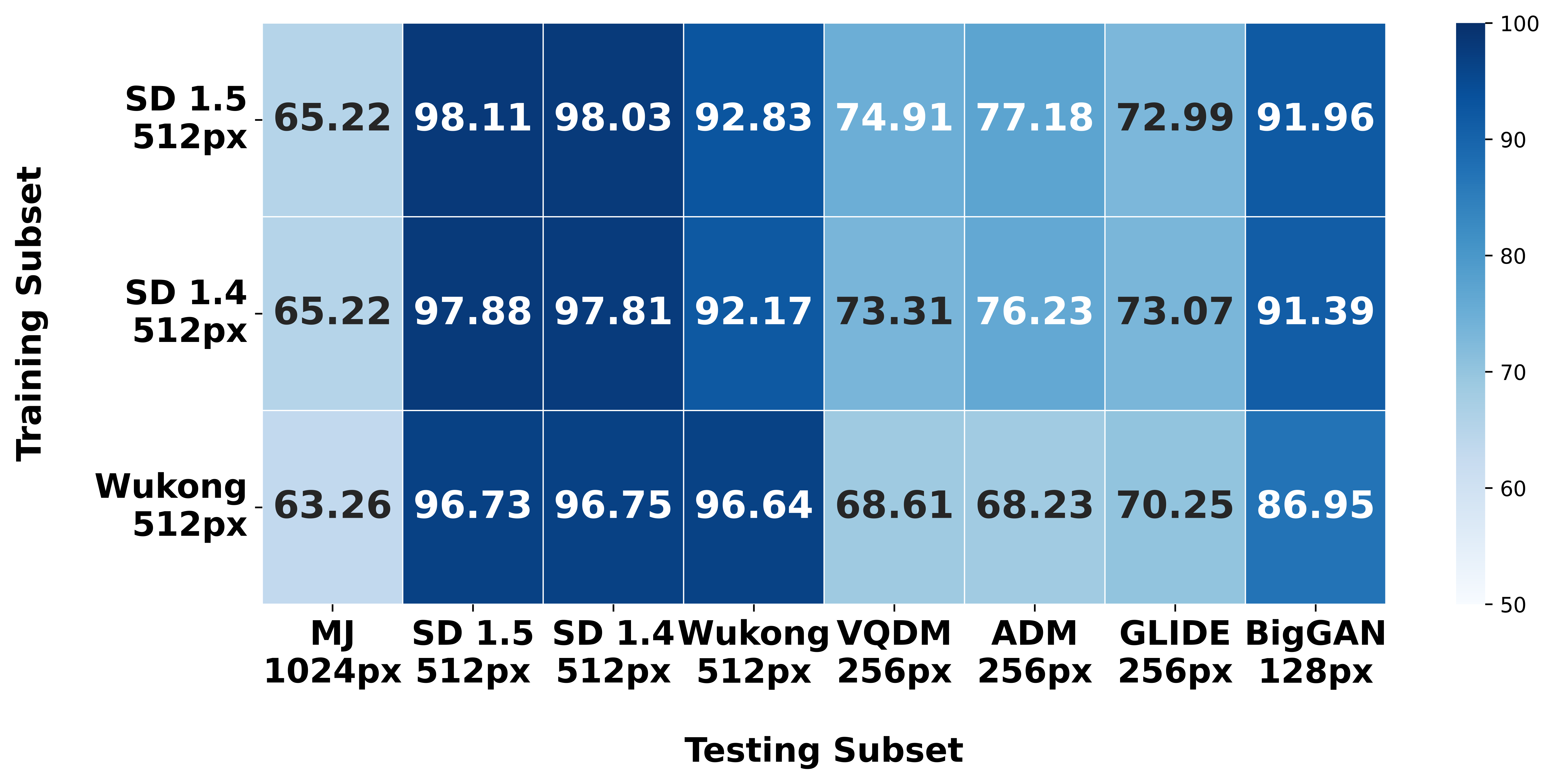}
        \caption{Accuracy (in \%) for size-constrained training on full validation datasets}
        \label{fig:controlledcrossperf}
    \end{subfigure}
    \hfill
    \begin{subfigure}[t]{0.45\textwidth}
        \centering
        \includegraphics[width=\linewidth]{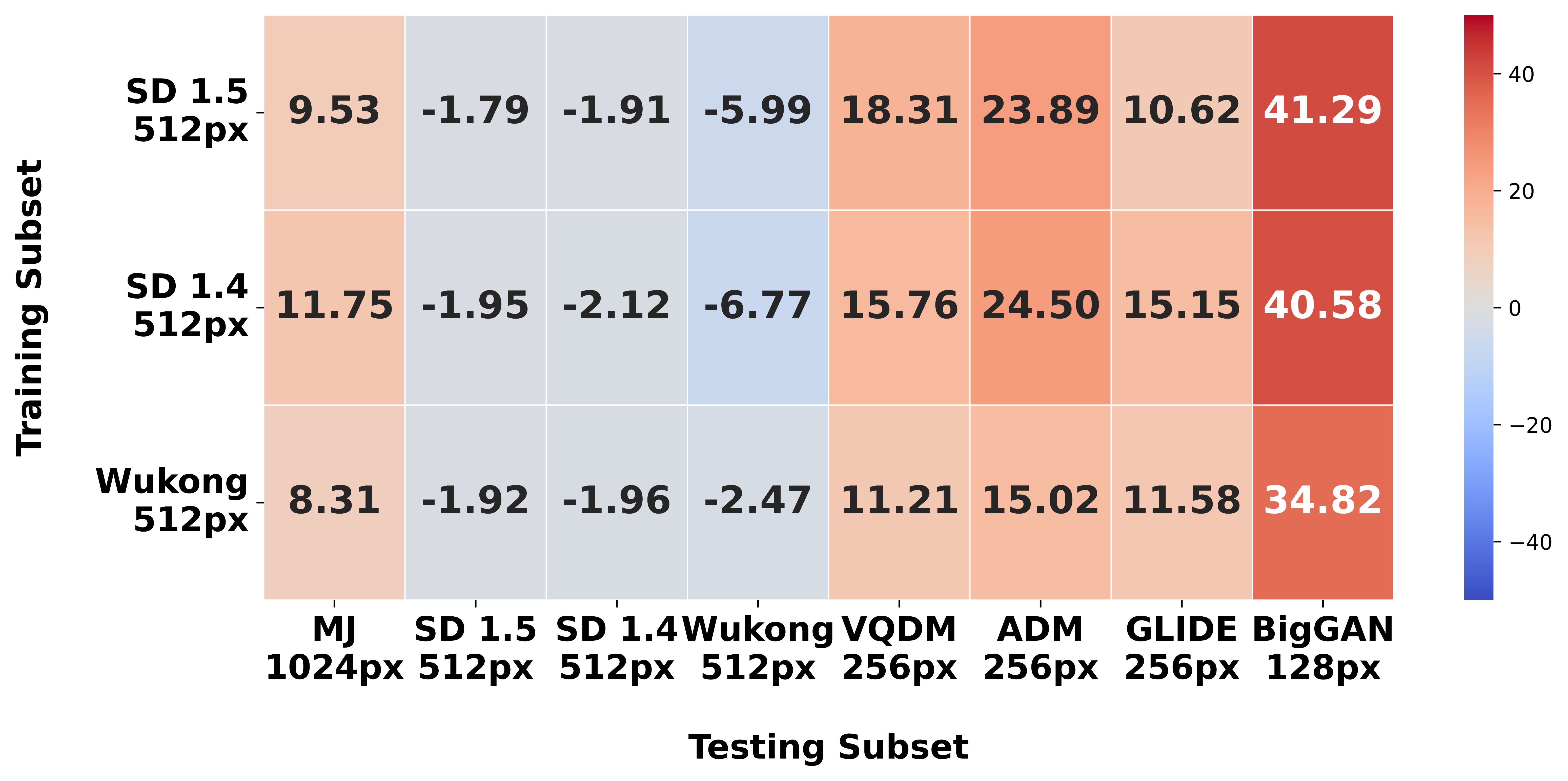}
        \caption{Difference to training on raw \textit{GenImage} dataset}
        \label{fig:controlleddiff}
    \end{subfigure}
    \caption{Cross-generator perfomance for \textit{ResNet50} detectors trained on compression and size constrained \textit{GenImage} training subset.}
    \label{fig:comparison}
    \setlength{\belowcaptionskip}{-30pt} % Adjust this value as needed
\end{figure}

\cref{fig:controlledcrossperf} shows the cross-generator performance of detectors trained with this size constraint, while \cref{fig:controlleddiff} highlights the difference compared to detectors trained on the unconstrained, raw dataset. Note that for the evaluation, images stored as \textit{PNG} files are compressed using a \textit{JPEG} quality factor of 96, ensuring the model is not exposed to uncompressed images, which it had not encountered during training. The results demonstrate an overall improvement in cross-generator generalization, with a maximum increase of 41.29 percentage points and an average increase of 11.06 percentage points, achieving state-of-the-art results in \textit{GenImage} cross-generator performance. We note a slight decrease in performance for subsets containing generated images of the same size as those used in training. This observation supports our hypothesis, since the original evaluation likely overestimated the generalization ability for generators of the same size: For instance, detectors trained on the stable-diffusion subset performed exceptionally well on the \textit{Wukong} subset, which shares the same size bias. Mitigating the size bias leads to a decline in generalization from stable-diffusion to \textit{Wukong} of approximately 7 percentage points. 

\subsection{Detector Ablation}
To ensure that the constraints not only improve the performance of \textit{ResNet50} detectors, we likewise examined our constrained training for a transformer based detector \textit{Swin-T}~\cite{liu2021swin}. ~\cref{tab:evaluation} summarizes the average scores: Note that we outperform the current stat-of-the-art, \textit{GenDet}~\cite{zhu2023gendet}, which only reports results of 81.6\% for training on \textit{SD1.4}. When validating our \textit{Swin-T} detectors, we achieve an average accuracy of 85.83\%. Specifically, the detector trained on the \textit{SD1.4} subset reaches an accuracy of 86.8\%. We provide the detailed cross-generator matrix of \textit{Swin-T} in \hyperref[app:c]{Appendix C}. 
The overall improvement in cross-generator performance is especially impressing when considering that the selection of training data reduced the number of samples by more than 75\%. Nevertheless, the key insight of our work is not merely enhanced performance and robustness but rather showing that the named biases significantly affect the detectors leading to a misjudgment when evaluating.

\begin{table}[t]
    \centering
    \caption{Average cross generator performance for \textit{ResNet50} and \textit{Swin-T} given in accuracy (in \%) when trained on raw \textit{GenImage} subsets and our constrained subsets.}
    \label{tab:evaluation}
    \small
    \begin{tabular}{ccccccc}
      \toprule
      \textbf{Training Subset} & \multicolumn{3}{c}{\textbf{ResNet50}} & \multicolumn{3}{c}{\textbf{Swin-T}} \\
      \cmidrule(lr){2-4} \cmidrule(lr){5-7}
       & Classic & Ours & Diff & Classic & Ours & Diff \\
      \midrule
      \textit{SD1.5} & 72.16 & 83.90 & +11.74 & 74.14 & 85.90 & +11.76 \\
      \textit{SD1.4} & 71.27 & 83.39 & +12.12 & 74.93 & 86.80 & +11.87 \\
      \textit{Wukong} & 71.61 & 80.93 & +9.32 & 73.20 & 84.80 & +11.60 \\
      \bottomrule
      Total & 71.68 & 82.74 & +11.06 & 74.09 & 85.83 & +11.74 \\
    \end{tabular}
  \end{table}

\section{Discussion and Conclusion}

Our findings demonstrate that datasets used for AI-generated image detection exhibit biases in compression artifacts and image dimensions. These biases hinder models from learning the core task of detecting generation specific artifacts and result in misaligned evaluations. By imposing constraints on the training dataset to mitigate these biases, we observed a significant shift in the evaluation of \textit{ResNet50} and \textit{Swin-T} detectors, yielding substantially improved robustness and generalization.
Since we only inspected these baseline methods, it would be interesting for future works to analyze if we can improve the performance of detection methods specifically designed for AI-generated image detection even more or whether they expose as only being able to perform well on biased datasets. While computational and time limitations prevented our evaluation of the \textit{DIRE} method yet, it remains a promising candidate for further investigation.

Nevertheless, it’s important to highlight that even with these constraints, datasets may contain other undesirable biases. A notable issue is the disparity in the source of natural images used for training generative models compared to those used for detector models, as shown in \cref{tab:datasets}. For instance, in the \textit{GenImage} dataset, \textit{Stable Diffusion} generators are trained on images from \textit{LAION}, whereas detector models use images from \textit{ImageNet}. Despite efforts to align image content through text prompts corresponding to \textit{ImageNet} classes, there is a potential risk that detectors might learn to distinguish between the styles of \textit{LAION} and \textit{ImageNet} images.
Even more concerning, some generative models like \textit{Stable Diffusion} actively fingerprint the generated content~\cite{kim2023wouaf}, leading to a undesirable distinction from the natural images of the detector that is clearly not transferable to other generative models. 

To create truly unbiased datasets, we propose that the detector models should be trained on the same natural images as their corresponding generative model. Additionally, natural and generated images should have near equal distributions in both compression and image dimensions.

\clearpage
\bibliographystyle{unsrt}
\bibliography{references}  %%% Uncomment this line and comment out the ``thebibliography'' section below to use the external .bib file (using bibtex) .

\begin{thebibliography}{10}

\bibitem{lu2023seeing}
Zeyu Lu, Di~Huang, Lei Bai, Jingjing Qu, Chengyue Wu, Xihui Liu, and Wanli
  Ouyang.
\newblock Seeing is not always believing: Benchmarking human and model
  perception of ai-generated images.
\newblock {\em Advances in Neural Information Processing Systems}, 36, 2024.

\bibitem{goodfellow2014generative}
Ian Goodfellow, Jean Pouget-Abadie, Mehdi Mirza, Bing Xu, David Warde-Farley,
  Sherjil Ozair, Aaron Courville, and Yoshua Bengio.
\newblock Generative adversarial nets.
\newblock {\em Advances in neural information processing systems}, 27, 2014.

\bibitem{zhang2019detecting}
Xu~Zhang, Svebor Karaman, and Shih-Fu Chang.
\newblock Detecting and simulating artifacts in gan fake images.
\newblock In {\em 2019 IEEE international workshop on information forensics and
  security (WIFS)}, pages 1--6. IEEE, 2019.

\bibitem{durall2020watch}
Ricard Durall, Margret Keuper, and Janis Keuper.
\newblock Watch your up-convolution: Cnn based generative deep neural networks
  are failing to reproduce spectral distributions.
\newblock In {\em Proceedings of the IEEE/CVF conference on computer vision and
  pattern recognition}, pages 7890--7899, 2020.

\bibitem{ho2020denoising}
Jonathan Ho, Ajay Jain, and Pieter Abbeel.
\newblock Denoising diffusion probabilistic models.
\newblock {\em Advances in neural information processing systems},
  33:6840--6851, 2020.

\bibitem{ricker2024detection}
Jonas Ricker, Simon Damm, Thorsten Holz, and Asja Fischer.
\newblock Towards the detection of diffusion model deepfakes.
\newblock {\em arXiv preprint arXiv:2210.14571}, 2022.

\bibitem{corvi2023intriguing}
Riccardo Corvi, Davide Cozzolino, Giovanni Poggi, Koki Nagano, and Luisa
  Verdoliva.
\newblock Intriguing properties of synthetic images: from generative
  adversarial networks to diffusion models.
\newblock In {\em Proceedings of the IEEE/CVF Conference on Computer Vision and
  Pattern Recognition}, pages 973--982, 2023.

\bibitem{ojha2023universal}
Utkarsh Ojha, Yuheng Li, and Yong~Jae Lee.
\newblock Towards universal fake image detectors that generalize across
  generative models.
\newblock In {\em Proceedings of the IEEE/CVF Conference on Computer Vision and
  Pattern Recognition}, pages 24480--24489, 2023.

\bibitem{zhu2023genimage}
Mingjian Zhu, Hanting Chen, Qiangyu Yan, Xudong Huang, Guanyu Lin, Wei Li,
  Zhijun Tu, Hailin Hu, Jie Hu, and Yunhe Wang.
\newblock Genimage: A million-scale benchmark for detecting ai-generated image.
\newblock {\em Advances in Neural Information Processing Systems}, 36, 2024.

\bibitem{zhu2023gendet}
Mingjian Zhu, Hanting Chen, Mouxiao Huang, Wei Li, Hailin Hu, Jie Hu, and Yunhe
  Wang.
\newblock Gendet: Towards good generalizations for ai-generated image
  detection.
\newblock {\em arXiv preprint arXiv:2312.08880}, 2023.

\bibitem{5206848}
Jia Deng, Wei Dong, Richard Socher, Li-Jia Li, Kai Li, and Li~Fei-Fei.
\newblock Imagenet: A large-scale hierarchical image database.
\newblock In {\em 2009 IEEE Conference on Computer Vision and Pattern
  Recognition}, pages 248--255, 2009.

\bibitem{wang2020cnngenerated}
Sheng-Yu Wang, Oliver Wang, Richard Zhang, Andrew Owens, and Alexei~A Efros.
\newblock Cnn-generated images are surprisingly easy to spot... for now.
\newblock In {\em Proceedings of the IEEE/CVF conference on computer vision and
  pattern recognition}, pages 8695--8704, 2020.

\bibitem{wang2023dire}
Zhendong Wang, Jianmin Bao, Wengang Zhou, Weilun Wang, Hezhen Hu, Hong Chen,
  and Houqiang Li.
\newblock Dire for diffusion-generated image detection.
\newblock {\em arXiv preprint arXiv:2303.09295}, 2023.

\bibitem{sarkar2023shadows}
Ayush Sarkar, Hanlin Mai, Amitabh Mahapatra, Svetlana Lazebnik, David~A
  Forsyth, and Anand Bhattad.
\newblock Shadows don't lie and lines can't bend! generative models don't know
  projective geometry... for now.
\newblock {\em arXiv preprint arXiv:2311.17138}, 2023.

\bibitem{he2015deep}
Kaiming He, Xiangyu Zhang, Shaoqing Ren, and Jian Sun.
\newblock Deep residual learning for image recognition.
\newblock In {\em Proceedings of the IEEE conference on computer vision and
  pattern recognition}, pages 770--778, 2016.

\bibitem{gragnaniello2021gan}
Diego Gragnaniello, Davide Cozzolino, Francesco Marra, Giovanni Poggi, and
  Luisa Verdoliva.
\newblock Are gan generated images easy to detect? a critical analysis of the
  state-of-the-art.
\newblock In {\em 2021 IEEE international conference on multimedia and expo
  (ICME)}, pages 1--6. IEEE, 2021.

\bibitem{cozzolino2021universal}
Davide Cozzolino, Diego Gragnaniello, Giovanni Poggi, and Luisa Verdoliva.
\newblock Towards universal gan image detection.
\newblock In {\em 2021 International Conference on Visual Communications and
  Image Processing (VCIP)}, pages 1--5. IEEE, 2021.

\bibitem{sinitsa2023deep}
Sergey Sinitsa and Ohad Fried.
\newblock Deep image fingerprint: Towards low budget synthetic image detection
  and model lineage analysis.
\newblock In {\em Proceedings of the IEEE/CVF Winter Conference on Applications
  of Computer Vision}, pages 4067--4076, 2024.

\bibitem{midjourney2022}
{Midjourney}.
\newblock \url{https://www.midjourney.com}, 2022.

\bibitem{autoamtic1111-stable-diffusion-webui}
Stable~Diffusion WebUI.
\newblock \url{https://github.com/AUTOMATIC1111/stable-diffusion-webui}, 2022.

\bibitem{wukong_2022}
Wukong.
\newblock \url{https://xihe.mindspore.cn/modelzoo/wukong}, 2022.

\bibitem{gu2022vector}
Shuyang Gu, Dong Chen, Jianmin Bao, Fang Wen, Bo~Zhang, Dongdong Chen, Lu~Yuan,
  and Baining Guo.
\newblock Vector quantized diffusion model for text-to-image synthesis.
\newblock In {\em Proceedings of the IEEE/CVF Conference on Computer Vision and
  Pattern Recognition}, pages 10696--10706, 2022.

\bibitem{dhariwal2021diffusion}
Prafulla Dhariwal and Alexander Nichol.
\newblock Diffusion models beat gans on image synthesis.
\newblock {\em Advances in neural information processing systems},
  34:8780--8794, 2021.

\bibitem{nichol2022glide}
Alex Nichol, Prafulla Dhariwal, Aditya Ramesh, Pranav Shyam, Pamela Mishkin,
  Bob McGrew, Ilya Sutskever, and Mark Chen.
\newblock Glide: Towards photorealistic image generation and editing with
  text-guided diffusion models.
\newblock {\em arXiv preprint arXiv:2112.10741}, 2021.

\bibitem{brock2019large}
A~Brock, J~Donahue, and K~Simonyan.
\newblock Large scale gan training for high fidelity natural image synthesis.
  international conference on learning representations.
\newblock 2019.

\bibitem{epstein2023online}
David~C. Epstein, Ishan Jain, Oliver Wang, and Richard Zhang.
\newblock Online detection of ai-generated images.
\newblock In {\em ICCV DeepFake Analysis and Detection Workshop}, 2023.

\bibitem{compression-imagenet}
towardsdatascience.
\newblock Compression in the {ImageNet} dataset.
\newblock
  \url{https://towardsdatascience.com/compression-in-the-imagenet-dataset-34c56d14d463}.

\bibitem{karras2019stylebased}
Tero Karras, Samuli Laine, and Timo Aila.
\newblock A style-based generator architecture for generative adversarial
  networks.
\newblock In {\em Proceedings of the IEEE/CVF conference on computer vision and
  pattern recognition}, pages 4401--4410, 2019.

\bibitem{liu2021swin}
Ze~Liu, Yutong Lin, Yue Cao, Han Hu, Yixuan Wei, Zheng Zhang, Stephen Lin, and
  Baining Guo.
\newblock Swin transformer: Hierarchical vision transformer using shifted
  windows.
\newblock In {\em Proceedings of the IEEE/CVF international conference on
  computer vision}, pages 10012--10022, 2021.

\bibitem{kim2023wouaf}
Changhoon Kim, Kyle Min, Maitreya Patel, Sheng Cheng, and Yezhou Yang.
\newblock Wouaf: Weight modulation for user attribution and fingerprinting in
  text-to-image diffusion models.
\newblock {\em arXiv preprint arXiv:2306.04744}, 2023.

\end{thebibliography}




% 	\bibitem{keshet2016prediction}
% 	Keshet, Renato, Alina Maor, and George Kour.
% 	\newblock Prediction-Based, Prioritized Market-Share Insight Extraction.
% 	\newblock In {\em Advanced Data Mining and Applications (ADMA), 2016 12th International 
%                       Conference of}, pages 81--94,2016.

% \end{thebibliography}

\newpage
\appendix
\section*{Appendix  \label{app:overview}}
\begin{figure}[h]
    \centering
    \includegraphics[width=1\linewidth]{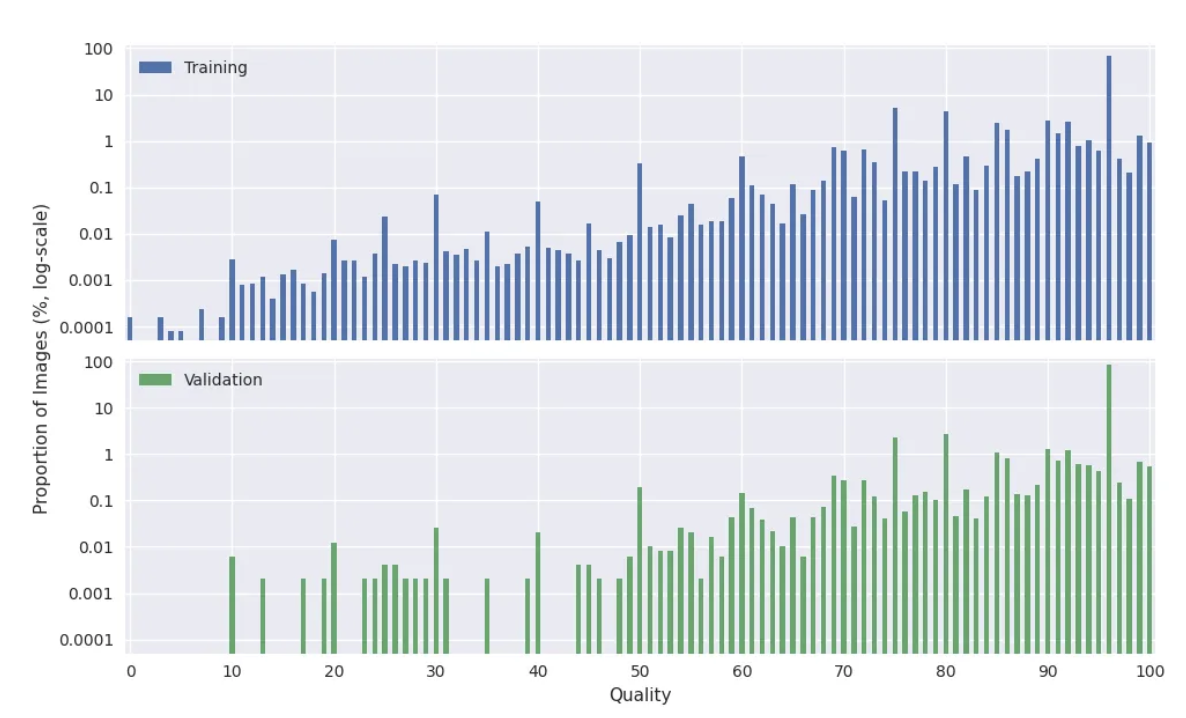}
    \caption{Distribution of \textit{JPEG} quality factor in \textit{ImageNet} with logarithmic scale. (Graphic from~\cite{compression-imagenet})}
    \label{fig:imagenet_qf_dist}
\end{figure}

\begin{figure}[h]
    \centering
    \includegraphics[width=1\linewidth]{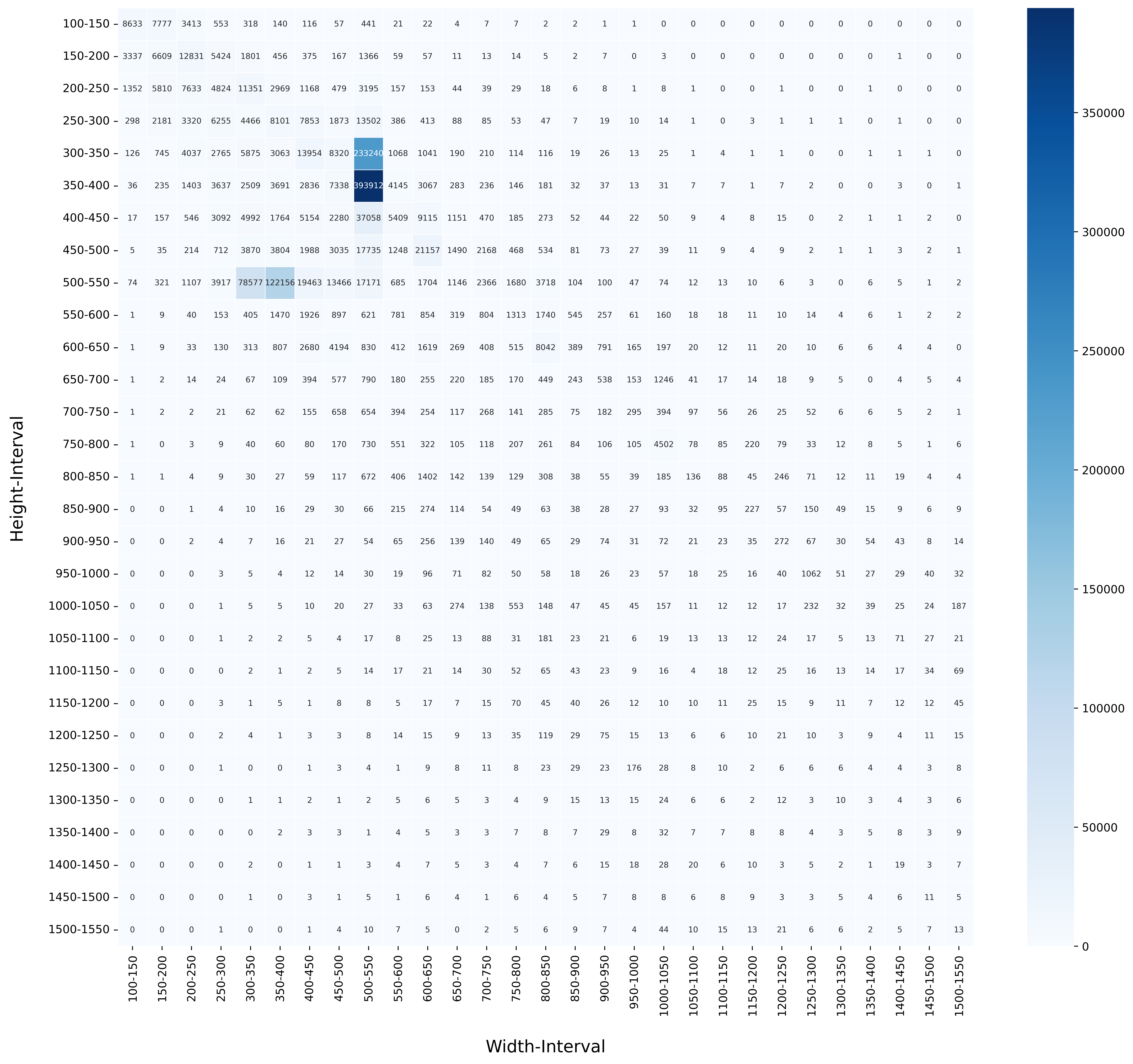}
    \caption{Distribution of image size in \textit{ImageNet} per interval.}
    \label{fig:imagenet_size_dist}
\end{figure}

\hyperref[app:a]{Appendix A} shows detailed insights into size (\cref{fig:imagenet_size_dist})and compression (\cref{fig:imagenet_qf_dist}) distribution in \textit{ImageNet} images, which are used in \textit{GenImage}. 
\hyperref[app:b]{Appendix B} shows the precision and recall corresponding to the experiment in \cref{sec:3.1} to show that compression leads to classifying an image as natural. 
\hyperref[app:c]{Appendix C} provides the detailed cross-generator matrix for the \textit{Swin-T} detector when trained on the constrained dataset and it's difference to training on the raw \textit{GenImage} dataset.
\section*{Appendix A \label{app:a}}

\clearpage

\section*{Appendix B \label{app:b}}

\begin{figure}[h]
    \centering
    \includegraphics[width=1\linewidth]{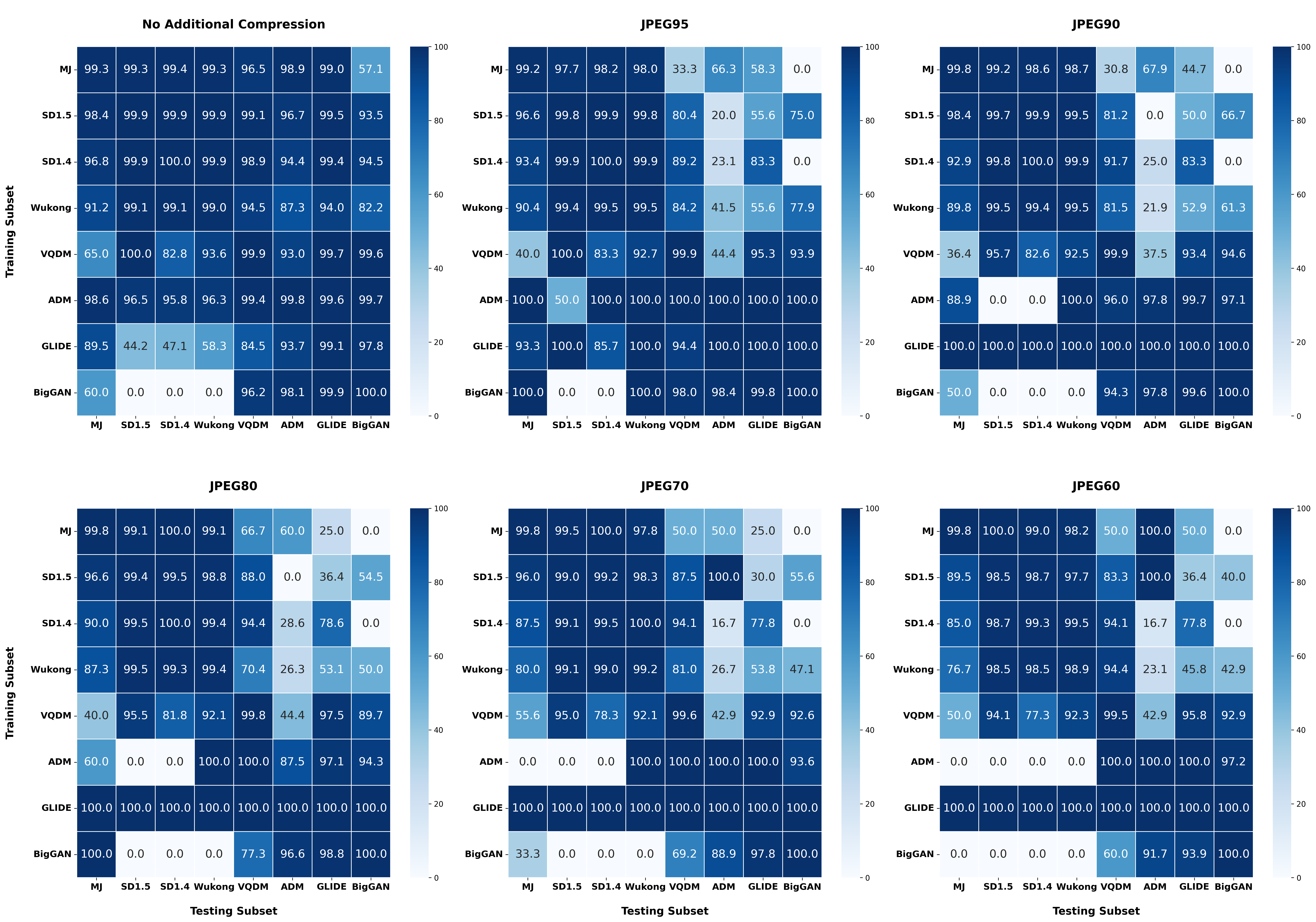}
    \caption{Precision (in \%) corresponding to \cref{fig:jpegcompression}. This demonstrates that the classification of natural images is not affected by the compression.}
    \label{fig:jpegcompression_precision}
\end{figure}

\begin{figure}[h]
    \centering
    \includegraphics[width=1\linewidth]{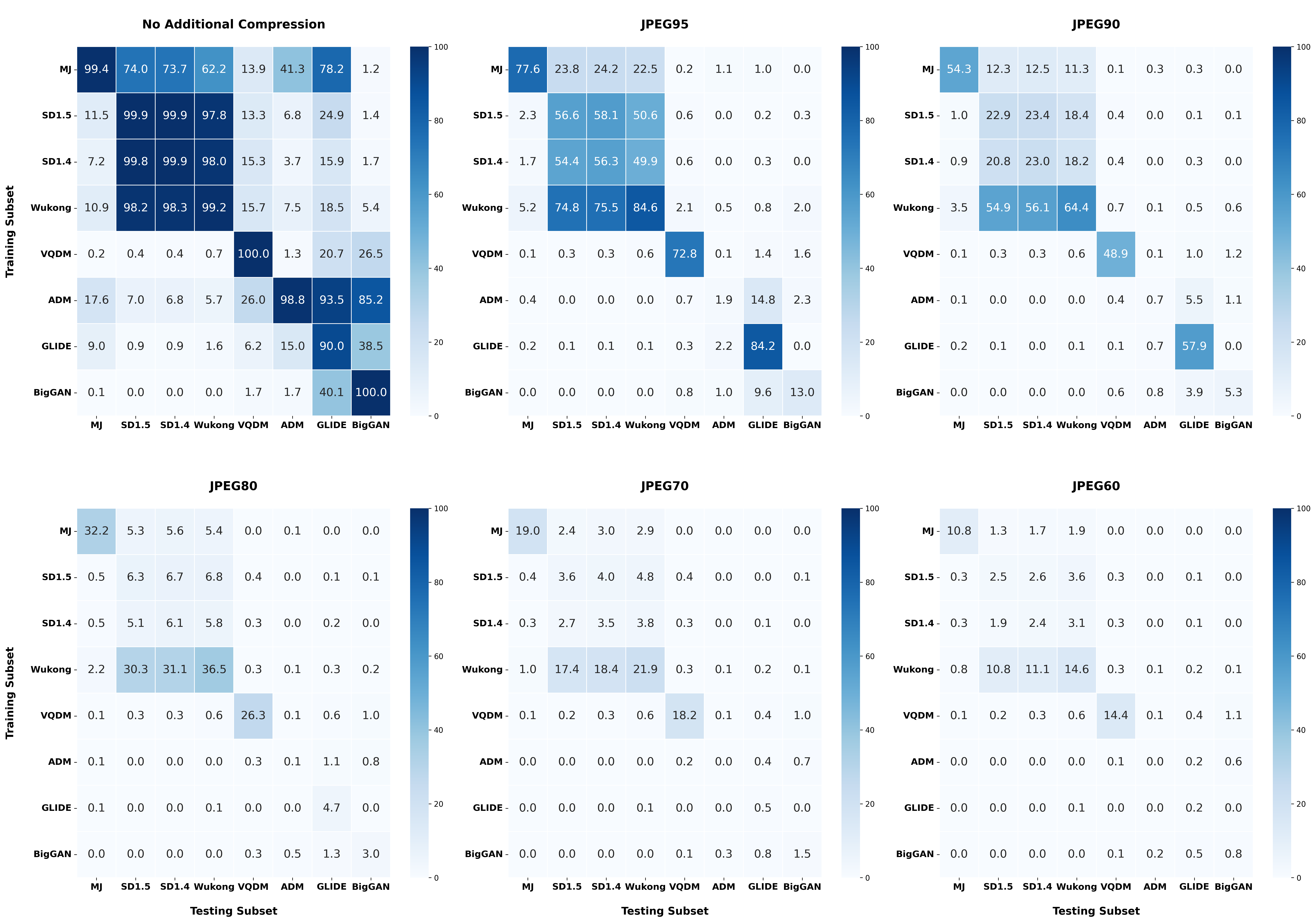}
    \caption{Recall (in \%) corresponding to \cref{fig:jpegcompression}. This demonstrates that compressed AI-generated images are likely classified as natural}
    \label{fig:jpegcompression_recall}
\end{figure}

\clearpage

\section*{Appendix C \label{app:c}}

\begin{figure}[h]
    \centering
    \begin{subfigure}[t]{0.45\textwidth}
        \centering
        \includegraphics[width=\linewidth]{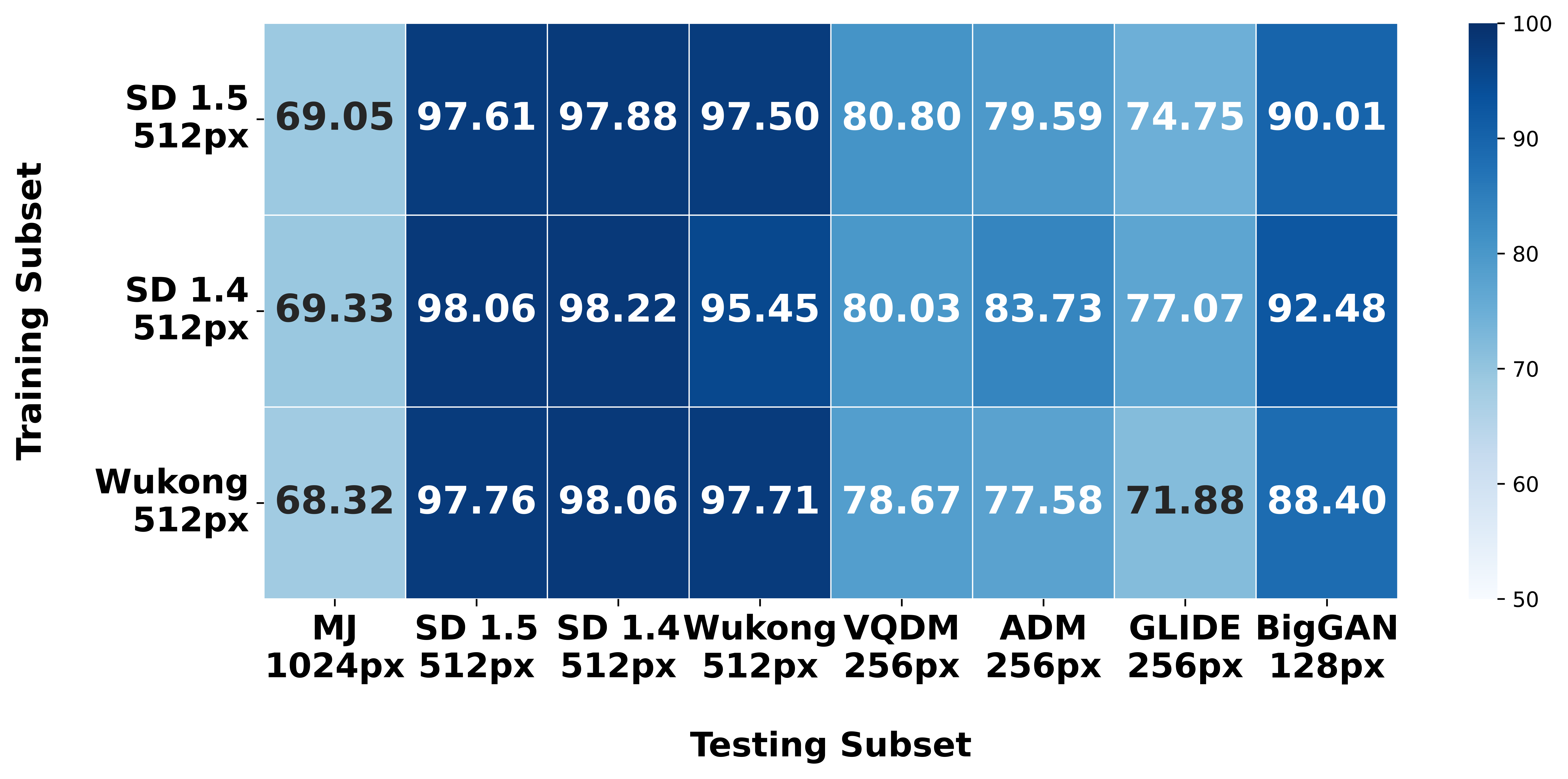}
        \caption{Accuracy (in \%) for constrained training in 4.2. on full validation datasets}
        \label{fig:controlledcrossperf_swin}
    \end{subfigure}
    \hfill
    \begin{subfigure}[t]{0.45\textwidth}
        \centering
        \includegraphics[width=\linewidth]{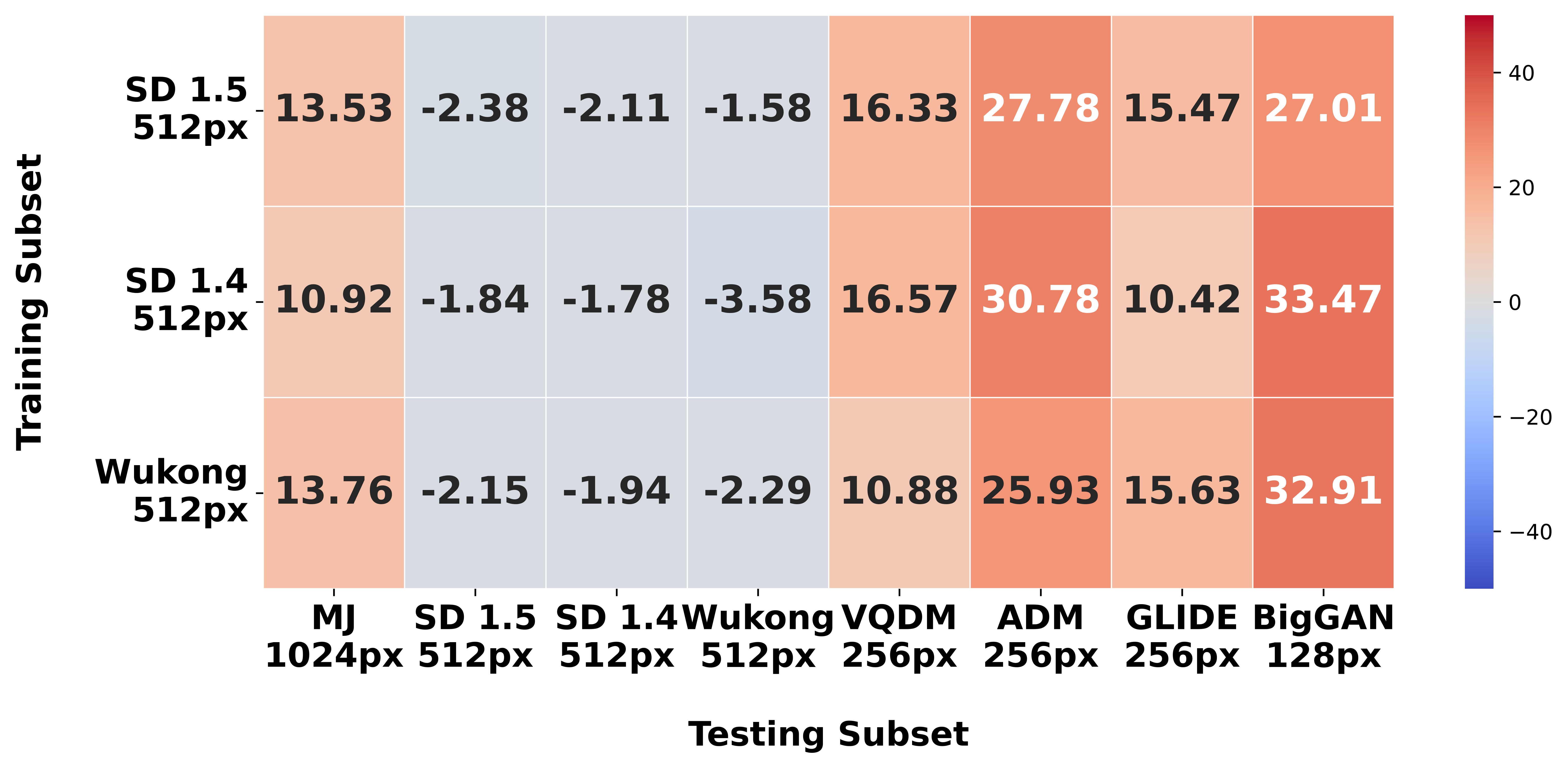}
        \caption{Difference to training on raw \textit{GenImage} dataset}
        \label{fig:controlleddiff_swin}
    \end{subfigure}
    \caption{Cross-generator perfomance for \textit{SWIN-T} detectors trained on compression and size constrained \textit{GenImage} training subset.}
    \label{fig:comparison_swin}
\end{figure}

\end{document}